\documentclass[sigconf,natbib=true]{acmart}
\AtBeginDocument{%
  }

\setcopyright{acmlicensed}
\acmDOI{XXXXXXX.XXXXXXX}

\usepackage{hyperref}
\usepackage{url}

\usepackage{algorithm}
\usepackage[noend]{algpseudocode}
\usepackage[utf8]{inputenc} 
\usepackage[T1]{fontenc}    
\usepackage{hyperref}       
\usepackage{url}            
\usepackage{booktabs}       
\usepackage{amsfonts}       
\usepackage{microtype}      
\usepackage{xcolor}         

\usepackage{amsmath} 
\usepackage{multirow}
\usepackage{tabularx}
\usepackage{booktabs}
\usepackage{graphicx} 
\usepackage{colortbl}
\usepackage{makecell}
\usepackage{arydshln}
\usepackage{wrapfig}
\usepackage{caption}
\usepackage{subcaption}
\usepackage{xcolor}
\definecolor{officeblue}{RGB}{79,129,189}   
\definecolor{officegreen}{RGB}{155,187,89}  
\definecolor{officeorange}{RGB}{217,150,148} 

\begin{document}
\def\method{MoRE}
\title{Mixture-of-Retrieval Experts for Reasoning-Guided Multimodal Knowledge Exploitation}%
\author{Chunyi Peng}
\authornote{ \ \ indicates equal contribution.}
\affiliation{%
  \institution{Northeastern University}
  \city{Shenyang}
  \country{China}
}
\email{pengchunyi@mails.neu.edu.cn}

\author{Zhipeng Xu}
\authornotemark[1]
\affiliation{%
  \institution{Northeastern University}
  \city{Shenyang}
  \country{China}
}
\email{xuzp@mails.neu.edu.cn}

\author{Zhenghao Liu}
\authornote{ \ \ indicates corresponding author.}
\affiliation{%
  \institution{Northeastern University}
  \city{Shenyang}
  \country{China}}
\email{liuzhenghao@mail.neu.edu.cn}

\author{Yishan Li}
\affiliation{%
  \institution{ModelBest. Inc}
  \city{Beijing}
  \country{China}
}
\email{liyishanthu@gmail.com}

\author{Yukun Yan}
\affiliation{%
  \institution{Tsinghua University}
  \city{Beijing}
  \country{China}
}
\email{yanyk.thu@gmail.com}

\author{Shuo Wang}
\affiliation{%
  \institution{Tsinghua University}
  \city{Beijing}
  \country{China}
}
\email{wangshuo.thu@gmail.com}

\author{Yu Gu}
\affiliation{%
  \institution{Northeastern University}
  \city{Shenyang}
  \country{China}}
\email{guyu@mail.neu.edu.cn}

\author{Minghe Yu}
\affiliation{%
  \institution{Northeastern University}
  \city{Shenyang}
  \country{China}}
\email{yuminghe@mail.neu.edu.cn}

\author{Ge Yu}
\affiliation{%
  \institution{Northeastern University}
  \city{Shenyang}
  \country{China}}
\email{yuge@mail.neu.edu.cn}

\author{Maosong Sun}
\affiliation{%
  \institution{Tsinghua University}
  \city{Beijing}
  \country{China}
}
\email{sms@tsinghua.edu.cn}

\renewcommand{\shortauthors}{Chunyi Peng et al.}

\begin{abstract}
Multimodal Retrieval-Augmented Generation (MRAG) has shown promise in mitigating hallucinations in Multimodal Large Language Models (MLLMs) by incorporating external knowledge. However, existing methods typically adhere to rigid retrieval paradigms by mimicking fixed retrieval trajectories and thus fail to fully exploit the knowledge of different retrieval experts through dynamic interaction based on the model's knowledge needs or evolving reasoning states. To overcome this limitation, we introduce \textbf{M}ixture-\textbf{o}f-\textbf{R}etrieval \textbf{E}xperts (\method{}), a novel framework that enables MLLMs to collaboratively interact with diverse retrieval experts for more effective knowledge exploitation. Specifically, \method{} learns to dynamically determine which expert to engage with, conditioned on the evolving reasoning state. To effectively train this capability, we propose Stepwise Group Relative Policy Optimization (Step-GRPO), which goes beyond sparse outcome-based supervision by encouraging MLLMs to interact with multiple retrieval experts and synthesize fine-grained rewards, thereby teaching the MLLM to fully coordinate all experts when answering a given query. Experimental results on diverse open-domain QA benchmarks demonstrate the effectiveness of \method{}, achieving average performance gains of over 7\% compared to competitive baselines. Notably, \method{} exhibits strong adaptability by dynamically coordinating heterogeneous experts to precisely locate relevant information, validating its capability for robust, reasoning-driven expert collaboration. All codes and data are released on \url{https://github.com/OpenBMB/MoRE}.
\end{abstract}

\maketitle

\section{Introduction}\label{sec:introduction}
Retrieval-Augmented Generation (RAG) methods~\cite{asai2023self, lewis2020retrieval, liu2025knowledge} empower Large Language Models (LLMs) to incorporate external information, thereby bridging the gap between their static parametric knowledge and dynamic external knowledge~\cite{bai2024hallucination, chen2024unified, zhao2023retrieving}. This integration has been shown to effectively mitigate hallucinations and improve response accuracy.
In practice, knowledge sources typically consist of multimodal information, spanning image-caption pairs, textual documents, and tabular data. Consequently, existing works often conduct separate evaluations across different modalities, such as Visual QA, Text QA, and Table QA, by predefining the modality to be retrieved for each task~\cite{abootorabi2025ask, xia2024rule, zhang2024mr}. However, in real-world scenarios, it is desirable to develop a universal RAG model capable of handling diverse tasks without relying on such strong prior assumptions regarding modality selection~\cite{yeo2025universalrag}.
\begin{figure}[t]
    \includegraphics[width=\linewidth]{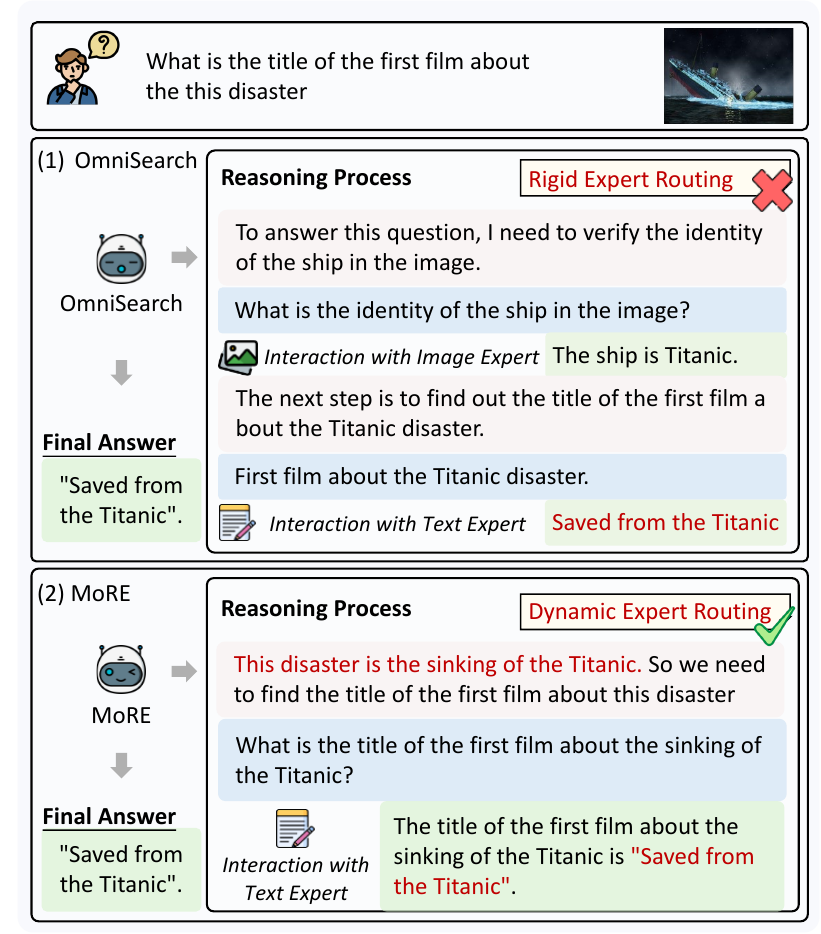}
    \caption{Illustration of Our Mixture-of-Retrieval Experts (\method{}) Model.\label{fig:intro}}
\end{figure}

To enable Multi-modal LLMs (MLLMs) to fully leverage knowledge from multiple modalities, existing methods~\cite{li2024benchmarking, yu2025unveiling, wu2025mmsearch} typically treat retrieval modules for different modalities as specialized experts, and employ the MLLM as a controller to conduct reasoning, plan expert selection, and generate intermediate observations based on the retrieved evidence.
Specifically, some approaches~\cite{yeo2025universalrag, yu2025unveiling} prompt the MLLM to decompose the input query into multiple sub-queries, determine the appropriate retrieval expert to interact with for each sub-query, and acquire modality-specific knowledge to answer the given query.
To better exploit knowledge from different retrieval experts, some works~\cite{li2024benchmarking} further optimize MLLMs to learn expert selection strategies by mimicking the retrieval expert interaction trajectories of teacher models through Supervised Fine-Tuning (SFT). However, this paradigm may lead to overfitting to fixed retrieval expert interaction patterns, rather than enabling the model to adaptively select experts tailored to the specific information needs during search. As shown in Figure~\ref{fig:intro}, the MLLM tends to follow a rigid retrieval pattern, first querying the image expert and then the text expert, without considering how the retrieved knowledge will actually be used in subsequent reasoning.
To address this limitation, the work~\cite{wu2025mmsearch} further optimizes MLLMs to identify more suitable retrieval paths using outcome-based reinforcement learning~\citep{shao2024deepseekmath}. While effective in jointly modeling retrieval and reasoning, such outcome-level supervision neglects the critical role of intermediate interactions with different retrieval experts and fails to explicitly integrate fine-grained interaction feedback from these experts to optimize the MLLM throughout the process of answering a given query.

In this paper, we introduce \method{}, a novel framework that designs a \textbf{M}ixture-\textbf{o}f-\textbf{R}etrieval \textbf{E}xperts mechanism to adaptively interact with diverse retrieval experts, thereby fully exploiting knowledge from different modalities.
To effectively optimize the capability of MLLMs in expert routing, we propose the Stepwise Group Relative Policy Optimization (Step-GRPO) method, which provides fine-grained rewards by integrating feedback from the interaction process with retrieval experts of different modalities throughout the entire reasoning trajectory, rather than relying solely on outcome-based rewards.
Specifically, \method{} leverages reasoning trajectories synthesized by a stronger MLLM as golden trajectories, and samples expert interaction trajectories at each step to obtain feedback on retrieval expert selection. These retrieval expert selection feedback signals are then used as supervision to optimize the MLLM.

Experimental results across diverse open-domain QA benchmarks validate the effectiveness of \method{}, yielding substantial performance gains of over 7\% compared to all baseline models. These results highlight the potential of \method{} as a unified framework for handling diverse tasks across multiple modalities. Moreover, \method{} demonstrates superior efficiency in leveraging multimodal experts by significantly reducing the number of required retrieval steps while maintaining high accuracy in Visual and Table QA tasks. This finding suggests that \method{} enables more effective planning and expert selection by incorporating feedback from retrieval expert interactions during optimization, rather than merely imitating the golden expert interaction trajectories.
\section{Related Work}
Existing RAG models~\cite{lewis2020retrieval, shi2023replug, ram2023context} usually focus on textual RAG modeling, where retrieved text documents are fed as contextual input to Large Language Models (LLMs)~\cite{ram2023context}.
Multimodal RAG (MRAG) methods~\cite{mei2025survey, sharifymoghaddam2024unirag, zhao2023retrieving, liu2025benchmarking} extend these textual RAG approaches by retrieving knowledge from sources of different modalities, thereby enhancing their capability to address more complex and diverse real-world applications~\cite{abootorabi2025ask, caffagni2024wiki, yu2024visrag, lahiri2024alzheimerrag}.
Typically, these methods predefine the modality for retrieval, search for relevant documents, and then ground the retrieved content into MLLMs to answer the given question~\cite{chen2022murag, wu2025visual}.
However, existing MRAG methods generally treat knowledge sources as static repositories and require predefining the golden knowledge sources to retrieve appropriate information for question answering. This design limits MRAG systems in actively leveraging knowledge across different modalities. Moreover, the reliance on predefined knowledge-base routing poses a major obstacle to building universal MRAG systems that can flexibly handle QA tasks involving diverse modalities~\cite{yeo2025universalrag, liu2025benchmarking, zhao2023retrieving}.

Instead of treating multimodal sources as static components, recent studies~\cite{yu2025unveiling, li2024benchmarking, yeo2025universalrag} conceptualize multimodal retrievers as distinct experts and explicitly coordinate their interactions to retrieve more task-specific knowledge. These approaches rely on the planning and reasoning capabilities of Multimodal Large Language Models (MLLMs) to route queries to appropriate retrieval experts and to acquire the necessary supporting evidence for answer generation.
For instance, UniversalRAG~\cite{yeo2025universalrag} introduces a modality-aware routing mechanism that dynamically identifies the most suitable modality for retrieval based on the input query and performs targeted search using the corresponding expert. OmniSearch~\cite{li2024benchmarking} adopts a modular pipeline that decomposes complex queries into sub-queries and assigns them to modality-specific experts, while CogPlanner~\cite{yu2025unveiling} leverages LLM-based agents to conduct reasoning-driven query decomposition and retrieval planning within iterative RAG workflows.
Although effective, these methods predominantly depend on prompting strategies or supervised fine-tuning to learn fixed expert routing behaviors, resulting in rigid retrieval schedules that are difficult to scale to more diverse and complex multimodal scenarios. Furthermore, the absence of explicit feedback signals regarding expert interactions prevents models from accurately estimating the utility of different experts or exploring alternative reasoning trajectories, thereby constraining their generalization beyond predefined routing policies~\cite{chu2025sft,yang2025r1}.

To better interact with these retrieval experts, recent work has applied Reinforcement Learning (RL)~\cite{kaelbling1996reinforcement} to optimize retrieval-augmented reasoning processes that focus on the text modality, with representative methods such as DPO~\cite{rafailov2023direct} and GRPO~\cite{shao2024deepseekmath}. DeepRAG~\cite{guan2025deeprag} combines a binary decision tree with DPO to train models to decide whether to retrieve or respond directly, guided by user preference signals. ReSearch~\cite{chen2025learning} unifies retrieval and reasoning within the GRPO framework by using final-answer accuracy as feedback, while R1-Searcher~\cite{song2025r1} adopts a two-stage, outcome-driven RL training strategy to enhance the search capability of LLMs. MMSearch-R1~\citep{wu2025mmsearch} further extends RL to multimodal RAG tasks by using outcome-based rewards to optimize MLLMs to reason about when and how to invoke different retrieval experts.
While effective at integrating retrieval with reasoning, these approaches primarily rely on outcome-based supervision and do not explicitly incorporate interaction feedback from different retrieval experts. 
\section{Methodology}\label{sec:method}
This section introduces the preliminaries of retrieval-augmented reasoning first (Sec.~\ref{sec:pre}), then describes the overall architecture of \method{} (Sec.~\ref{sec:overview}), and finally presents the stepwise reinforcement learning–based optimization strategy adopted by \method{} (Sec.~\ref{sec:stepgrpo}).

\begin{figure*}[t]
    \includegraphics[trim=0 40 0 0 ,clip,width=\linewidth]{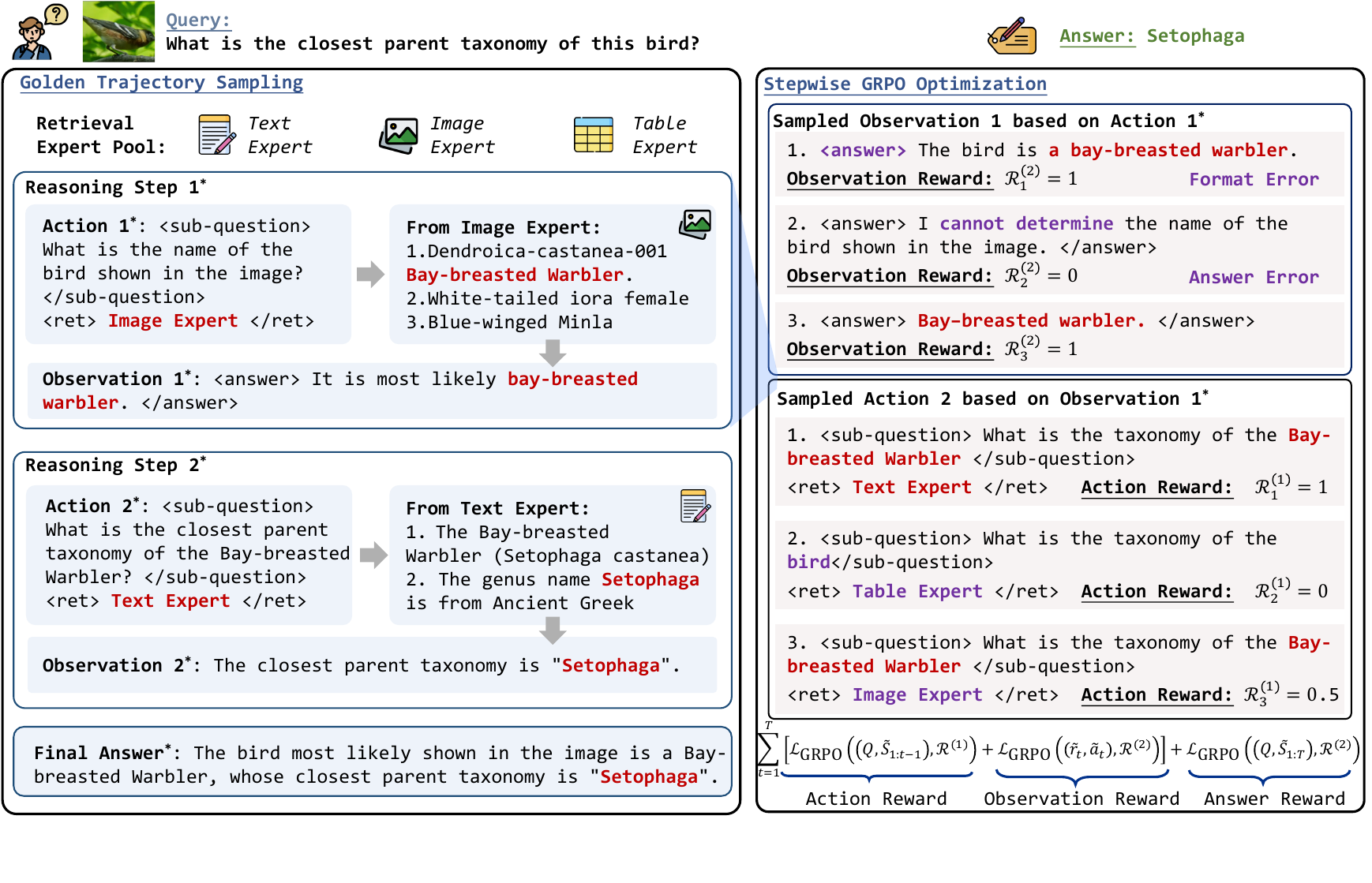}
    \caption{The Architecture of Stepwise GRPO Training Used by \method{}. }
    \label{fig:main}
\end{figure*} 
\subsection{Preliminaries of Retrieval-Augmented Reasoning}\label{sec:pre}
In this subsection, we first introduce background knowledge on retrieval during reasoning. Following existing works~\cite{li2025search}, retrieval is interleaved with explicit reasoning steps, where it is invoked on demand during inference to acquire the required information, and intermediate answers are progressively constructed throughout the retrieved knowledge and the evolving reasoning trajectory.

Given an input query $Q$ and a text corpus $\mathcal{D}_\text{text}$, we model the agent's reasoning process as a multi-step trajectory.
At each step $t$, the reasoning process $s_t$ follows a reasoning-action-observation paradigm:
\begin{equation}\label{eq:ra_reason_step}
    s_t = (r_t, a_t, o_t),
\end{equation}
where the agent first produces an explicit reasoning trace $r_t$ conditioned on previous reasoning trajectories:
\begin{equation}
r_t \sim \pi_\theta(\cdot \mid Q, s_1, \dots, s_{t-1}).
\end{equation}

This reasoning trace performs self-assessment over the current evidence, identifies missing information, and determines whether external retrieval is necessary, without directly interacting with the retrieval expert.
Conditioned on both previous reasoning trajectories and the generated reasoning $r_t$, the agent then decides a retrieval action:
\begin{equation}
a_t \sim \pi_\theta(\cdot \mid Q, s_1, \dots, s_{t-1}, r_t),
\end{equation}
where the action $a_t$ can be a search expert that contains a generated query $q_t$ to search related documents as the observations $o_t$:
\begin{equation}
o_t = \text{RetrievalExpert} (q_t, \mathcal{D}_\text{text}).
\end{equation}

Finally, conditioned on the complete $T$ reasoning steps $(s_1, \dots, s_T)$, the policy model generates the final answer $y$:
\begin{equation}
y \sim \pi_\theta(\cdot \mid s_1,\dots, s_T),
\end{equation}
where $T$ denotes the maximum of either the predefined maximum number of iterations or the step at which the model judges that the existing knowledge is sufficient.
Although existing works have demonstrated promising effectiveness, they are primarily confined to textual question answering tasks, thereby limiting the ability of agentic RAG models to autonomously acquire and leverage broader world knowledge~\cite{ge2023augmenting}. To address the problem, this paper proposes the Mixture-of-Retrieval Experts (\method{}) method to seek knowledge from knowledge sources of different modalities (Sec~\ref{sec:overview}).

\subsection{Mixture-of-Retrieval Experts for Multimodal Knowledge Collaboration}
\label{sec:overview}
To effectively integrate heterogeneous external knowledge, we reconceptualize multimodal retrieval as a structured mixture of modality-specialized retrieval experts. 

In practice, different sub-problems within a single query may require fundamentally distinct forms of evidence, such as conceptual knowledge for understanding, visual signals for perceptual grounding, or structured tables for precise numerical reasoning. However, in real-world scenarios, the modality of information is often unavailable as prior knowledge for RAG models to guide modality-specific retrieval.
To address this challenge, \method{} first models each modality-specific retriever as an independent expert, assembling three representative experts for mixture-of-retrieval experts modeling. \method{} then leverages MLLMs as a central reasoning controller, dynamically selecting which expert to invoke at each step to enable adaptive and fine-grained knowledge acquisition. This design supports more effective question answering through emergent multimodal collaboration. Additionally, \method{} optimizes the LLMs to perform more effective expert routing and knowledge retrieval through the optimization process described in Sec.~\ref{sec:stepgrpo}.

\textbf{Retrieval Expert Pool.}
We define a set of modality-specific retrieval experts as:
\begin{equation}
\mathcal{D} = \{\mathcal{D}_{\text{text}}, \mathcal{D}_{\text{image}}, \mathcal{D}_{\text{table}}\},
\end{equation}
where each expert is equipped with an independent indexing and retrieval pipeline tailored to its modality:

\begin{itemize}
    \item \texttt{Text Expert} $\mathcal{D}_{\text{text}}$ contains unstructured textual knowledge, providing entity descriptions, background facts, and explanatory evidence for long-tail or open-domain concepts.
    \item \texttt{Image Expert} $\mathcal{D}_{\text{image}}$ stores images or image-text pairs and supports visual grounding by extracting fine-grained attributes, spatial relations, and scene-level semantics.
    \item \texttt{Table Expert} $\mathcal{D}_{\text{table}}$ comprises structured data and supports accurate numerical lookup, comparison, and aggregation, addressing the known limitations of large language models in precise numerical reasoning.
\end{itemize}

\textbf{Reasoning-Driven Expert Selection.}
At each step $t$, the agent operates following a reasoning–action–observation paradigm $s_t = (r_t, a_t, o_t)$. In contrast to the traditional retrieval-augmented reasoning paradigm (Eq.~\ref{eq:ra_reason_step}), \method{} first determines which expert to select and then generates an expert-specific sub-query. Formally, a retrieval action is defined as:
\begin{equation}
a_t = (a^\text{select}_t, a^\text{search}_t),
\end{equation}
where $a^\text{select}_t$ denotes the expert routing action. The component $a^\text{search}_t$ contains a sub-query synthesized by the model conditioned on the reasoning $r_t$, or generates a ``NULL'' token to indicate that the current knowledge is sufficient.
Upon receiving a retrieval action, the environment employs the expert-specific retriever to search within the selected expert $\mathcal{D}(a_t^\text{select})$ using the generated query $a_t^\text{search}$, yielding the retrieved knowledge:
\begin{equation}
o_t = \text{RetrievalExpert}(a^\text{search}_t, \mathcal{D} (a^\text{select}_t)).
\end{equation}

\subsection{Learning Routing Policies for Retrieval Experts via Stepwise GRPO}
\label{sec:stepgrpo}
When mixing multiple retrieval experts, learning how to route queries and effectively leverage different experts is crucial for modeling \method{}. Thus, as shown in Figure~\ref{fig:main}, we propose a Stepwise Group Relative Policy Optimization (Step-GRPO) approach that optimizes MLLMs to interact more effectively with diverse retrieval experts when answering a given query. Specifically, we first sample several high-quality reasoning trajectories as supervision signals, then prompt the model to perform stepwise sampling to explore different expert interaction patterns. Finally, we jointly optimize expert selection and knowledge utilization through GRPO.

\textbf{Reward Modeling in \method{}.}
To optimize each reasoning process $s_t$, \method{} defines two types of rewards, action reward $\mathcal{R}^{(1)}$ and the observation reward $\mathcal{R}^{(2)}$, to guide the model in (1) routing to an appropriate retrieval expert to retrieve information and (2) producing a more accurate answer $a_i$ within the exploration trajectory.

First, $\mathcal{R}^{(1)}$ captures the fine-grained interaction feedback derived from retrieval experts at each reasoning step $s_t$ ($1 \leq t \leq T$). 
This reward functions as a dual-feedback mechanism to guide the retrieval process: 
(i) feedback on query quality $\mathcal{R}_\text{ask}(a_t^\text{search})$, which indicates whether the generated query is targeted and on-point by measuring its semantic similarity with the corresponding golden query in $s_t^*$ using the BGE-M3 embedding model~\cite{chen2024bge}; 
and (ii) feedback on expert selection  $\mathcal{R}_\text{route}(a_t^\text{select})$, which validates whether the model has correctly routed the query to the appropriate retrieval expert. 
The overall action reward $\mathcal{R}^{(1)}$ is defined as:
\begin{equation}\label{eq:reward1}\small
    \mathcal{R}^{(1)} = \mathcal{R}_\text{format}(a_t) \times(\alpha \mathcal{R}_\text{ask}(a_t^\text{search})+\beta \mathcal{R}_\text{route}(a_t^\text{select})),
\end{equation}
where $\alpha$ and $\beta$ are hyperparameters balancing the importance of query relevance and routing correctness. 
The formatting reward $\mathcal{R}_\text{format}(a_t)$ ensures that both the query and the selected retrieval expert are strictly enclosed in the required special tokens.


Second, to optimize the observation $o_t$ at each step $s_t$, we define the following reward:
\begin{equation}\label{eq:reward2}\small
    \mathcal{R}^{(2)} = \mathcal{R}_\text{format}(o_t) \times \mathcal{R}_\text{answer}(o_t),
\end{equation}
where $\mathcal{R}_\text{answer}(o_t)$ evaluates whether the generated observation $o_t$ is correct. We assess the observation quality by using accuracy and F1-Recall~\cite{li2024benchmarking}, where F1-Recall is applied to intermediate observations $o_t$ ($1 \leq t \leq T$), which often correspond to long and complex MLLM-generated references, and for the final answer, $y$, we use the accuracy for evaluation, given the short golden reference. The formatting reward $\mathcal{R}_\text{format}(o_t)$ enforces that the answer is enclosed in the special token.

\textbf{Step-GRPO Training.}
To enable effective learning of routing policies without relying on costly manual annotations, we first construct a collection of reasoning trajectories by leveraging multiple models for trajectory synthesis and then select the golden trajectories for training. Based on these trajectories, we further adopt a stepwise sampling strategy to conduct Step-GRPO training.

Specifically, we employ a prompt-guided sampling strategy, in which the model is prompted with a set of few-shot examples to generate candidate trajectories for each training query. We then introduce a rigorous dual-filtering mechanism to identify the golden trajectory $\Tilde{S}$, ensuring that the ground-truth trajectory $\Tilde{S}$ successfully leads to the ground-truth answer $\Tilde{y}$:
\begin{equation}
\Tilde{S} = \left[ \Tilde{s}_1, \Tilde{s}_2, \dots, \Tilde{s}_T \right],
\end{equation}
where $\Tilde{s}_t = (\Tilde{r}_t, \Tilde{a}_t, \Tilde{o}_t)$ denotes the $t$-th step of the golden trajectory $\Tilde{S}$. Specifically, $\Tilde{r}_t$, $\Tilde{a}_t$, and $\Tilde{o}_t$ represent the golden reasoning, action, and observation results at the $t$-th step, respectively. The action $\Tilde{a}_t$ consists of both the golden expert selection action $\Tilde{a}^\text{select}_t$ and the golden search action $\Tilde{a}^\text{search}_t$.

Based on the obtained golden trajectory $\Tilde{S}$, we further design a stepwise sampling scheme to optimize the model's capabilities in expert selection, search action, and knowledge utilization. This optimization is achieved by minimizing the following loss:
\begin{equation}\label{eq:stepgrpo}\small 
\begin{aligned}
    \mathcal{L} = \sum_{t=1}^{T}[\mathcal{L}_\text{GRPO}((Q, \Tilde{S}_{1:t-1}), \mathcal{R}^{(1)}) + \mathcal{L}_\text{GRPO}((\Tilde{r}_t, \Tilde{a}_t), \mathcal{R}^{(2)})]\\
+ \mathcal{L}_\text{GRPO}((Q, \Tilde{S}_{1:T}), \mathcal{R}^{(2)}),
\end{aligned}
\end{equation}
where $\mathcal{L}_\text{GRPO}((Q, \Tilde{S}_{1:T}), \mathcal{R}^{(2)})$ denotes the loss for generating the final answer conditioned on the given query $Q$ and the complete reasoning trajectory $\Tilde{S}_{1:T}$. At the $t$-th step, $\mathcal{L}_\text{GRPO}((Q, \Tilde{S}_{1:t-1}), \mathcal{R}^{(1)})$ and $\mathcal{L}_\text{GRPO}((\Tilde{r}_t, \Tilde{a}_t), \mathcal{R}^{(2)})$ are used to optimize expert selection and search actions, respectively.
Each GRPO loss term $\mathcal{L}_{\text{GRPO}}(x, \mathcal{R})$ is computed over a batch of trajectories sampled from the old policy model $\pi_{\theta_{\text{old}}}$, given an input $x$ and reward signal $\mathcal{R}$:
\begin{equation}\small
\begin{aligned}
\mathcal{L}_{\text{GRPO}}(x, \mathcal{R}) = 
&- \frac{1}{G} \sum_{k=1}^{G} 
\min \Bigg(
\frac{\pi_\theta(\mathcal{O}_k \mid x)}{\pi_{\theta_{\text{old}}}(\mathcal{O}_k \mid x)} \hat{A}_k (\mathcal{R}), \\
& \text{clip}\Big(
\frac{\pi_\theta(\mathcal{O}_k \mid x)}{\pi_{\theta_{\text{old}}}(\mathcal{O}_k \mid x)},
1-\epsilon, 1+\epsilon
\Big)\hat{A}_k (\mathcal{R})
\Bigg),
\end{aligned}
\end{equation}
where $\epsilon$ is the clipping hyperparameter and $\pi_\theta$ denotes the current policy model. Each $\mathcal{O}_k$ corresponds to a token sequence sampled from $\pi_{\theta_{\text{old}}}$.
For a given input $x$, we sample a group of responses $\{\mathcal{O}_1, \mathcal{O}_2,\dots, \mathcal{O}_G\}$, and use the interaction feedback $\mathcal{R}^{(1)}$ or $\mathcal{R}^{(2)}$ to calculate their corresponding rewards $\{\mathcal{R}(\mathcal{O}_1), \mathcal{R}(\mathcal{O}_2), \dots, \mathcal{R}(\mathcal{O}_G)\}$. The normalized advantage estimate $\hat{A}_k (\mathcal{R})$ for each token is computed as:
\begin{equation}\label{eq:advatage}\small
    \hat{A}_k (\mathcal{R}) = \frac{r_k-\text{mean}(\{\mathcal{R}(\mathcal{O}_1), \mathcal{R}(\mathcal{O}_2),\dots,\mathcal{R}(\mathcal{O}_G)\})}{\text{std}(\{\mathcal{R}(\mathcal{O}_1), \mathcal{R}(\mathcal{O}_2),\dots,\mathcal{R}(\mathcal{O}_G)\})}.
\end{equation}

\section{Experimental Methodology}

\begin{table}[t]
 \centering
 \caption{Data Statistics.} 
 \label{tab:datas}
 \begin{tabular}{lllr}\toprule
    \multirow{1}{*}{\textbf{Split}} & \multirow{1}{*}{\textbf{Scenarios}} 
     & \multirow{1}{*}{\textbf{Dataset}} & \multicolumn{1}{c}{\textbf{Total}}  \\ 
     \midrule
     \multirow{3}{*}{Training} & TextQA & 2WikiMultihopQA~\cite{ho20202wikimultihopqa} & 500 \\
     & VQA & InfoSeek~\cite{chen2023can} & 1,000  \\
     & TableQA & Open-WikiTable~\cite{kweon2023open} & 500  \\\midrule
     \multirow{6}{*}{Evaluation} & TextQA & 2WikiMultihopQA~\cite{ho20202wikimultihopqa} & 1,000   \\
     & VQA & InfoSeek~\cite{chen2023can} & 1,000   \\
     & VQA & Dyn-VQA~\cite{li2024benchmarking} & 715  \\
     & VQA & WebQA~\cite{chang2022webqa} & 1,000 \\
     & TableQA & Open-WikiTable~\cite{kweon2023open} & 1,000  \\
     & TableQA & TabFact~\cite{chen2019tabfact} & 1,000   \\
    \bottomrule
 \end{tabular}
\end{table}

\begin{figure}[!h]
    \includegraphics[width=\linewidth]{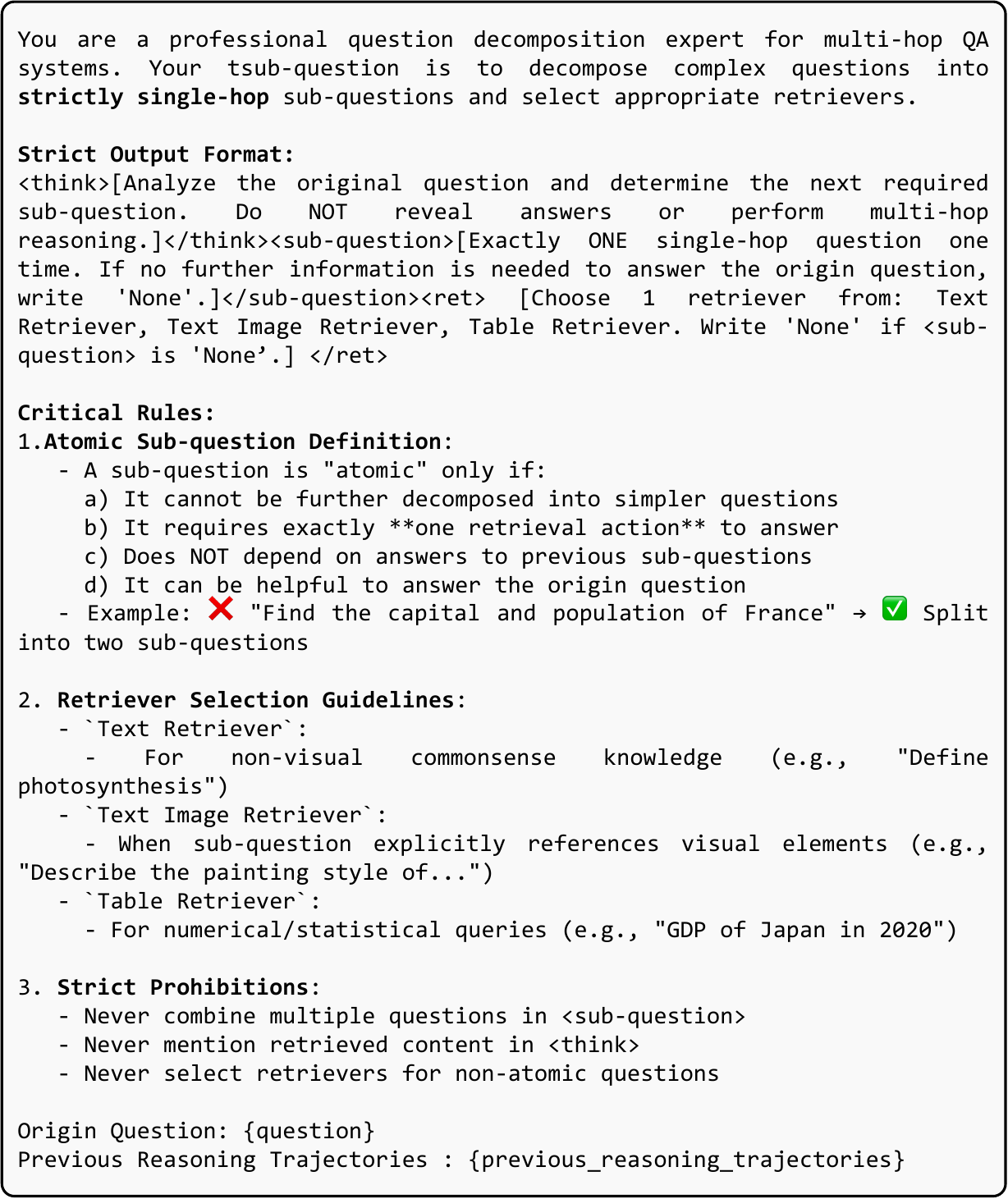}
    
    \caption{The Prompt Template of \method{} (Step-GRPO) for Query Generation and Retriever Selection.}
    \label{fig:prompt_router}
\end{figure}
This section describes the datasets, retrieval expert settings, evaluation metrics, baselines, and implementation details. 

\textbf{Datasets.} 
We evaluate our framework on a diverse set of open-domain QA benchmarks that require complex reasoning across textual (2WikiMultihopQA~\cite{ho20202wikimultihopqa}), visual (InfoSeek~\cite{chen2023can}, Dyn-VQA~\cite{li2024benchmarking}, and \texttt{WebQA}~\cite{chang2022webqa}), and tabular (Open-WikiTable~\cite{kweon2023open} and TabFact~\cite{chen2019tabfact}) modalities. To strictly assess the generalization capability of \method{}, we adopt a cross-dataset evaluation setting. Specifically, we utilize 2WikiMultihopQA for Text QA, InfoSeek for Visual QA, and Open-WikiTable for Table QA as the training sets.
For evaluation, we employ WebQA, Dyn-VQA, and TabFact as out-of-distribution test sets. WebQA and Dyn-VQA require multi-hop reasoning over heterogeneous retrieval experts, while TabFact challenges the model's ability to verify facts against structured tabular data. Detailed data statistics can be seen in Table~\ref{tab:datas}.

\textbf{Retrieval Experts.}
In our framework, we conceptualize each domain-specific corpus as an independent expert, equipped with a specialized retriever to access its knowledge. For the \texttt{Text Expert}, we employ the English Wikipedia dump-\texttt{20241020}\footnote{\url{https://dumps.wikimedia.org}} as the knowledge source, segmented into approximately 52 million passage-level retrieval units, and utilize BGE-M3~\citep{chen2024bge} as the corresponding text retriever. For the \texttt{Image Expert}, we utilize the M-BEIR corpus~\citep{wei2024uniir}, which contains over 5.6 million text-image pairs, and use UniIR~\citep{wei2024uniir} as the text-image retriever. We adopt the Open-WikiTable corpus~\citep{kweon2023open}, consisting of 24,680 structured tables, and employ a specialized dense table retriever~\citep{kweon2023open} following the Open-WikiTable setup to locate relevant table segments for numerical and statistical reasoning. 

\textbf{Evaluation Metrics.} Following prior work~\cite{li2024benchmarking}, our experiments utilize the F1-Recall score for evaluation.

\textbf{Baselines.}
We compare \method{} against three categories of baselines: vanilla (M)LLMs, vanilla RAG methods, and RAG methods with expert routing, to evaluate its expert routing and reasoning capabilities.
For vanilla (M)LLMs, models are prompted to generate answers relying solely on their internal parametric knowledge, without accessing any external retrieval experts. We evaluate Qwen2.5-VL-7B~\citep{bai2025qwen2} and R1-Distill-Qwen-32B~\citep{guo2025deepseek}, where both models are prompted to answer queries without access to any external knowledge.
To compare with standard RAG approaches, we include Vanilla RAG, which retrieves the top-$k$ documents from a pre-selected source; IRCOT~\citep{trivedi2022interleaving} and IterRetGen~\citep{shao2023enhancing}, which perform iterative retrieval to acquire sufficient knowledge; and Search-O1~\citep{li2025search}, which leverages reasoning capabilities to plan, search, and answer the given query. Moreover, we configure these baselines with both single-modality experts (utilizing only the text, image, or table expert individually) and a comprehensive expert ensemble (simultaneously utilizing all three experts) to provide external knowledge. 
We also compare our approach with RAG methods equipped with expert routing mechanisms. Specifically, we include CogPlanner~\citep{yu2025unveiling} and UniversalRAG~\citep{yeo2025universalrag}, which primarily rely on prompts to guide the model in generating queries to select appropriate retrieval experts. OmniSearch~\citep{li2024benchmarking} is also compared in our experiments, which focuses on using SFT to mimic the routing trajectories and reasoning patterns of teacher models. Additionally, we compare our approach with MMSearch-R1~\citep{wu2025mmsearch}, which incorporates reinforcement learning to optimize retrieval and reasoning capabilities based on outcome-based rewards.

\begin{figure}[!t]
    \includegraphics[width=\linewidth]{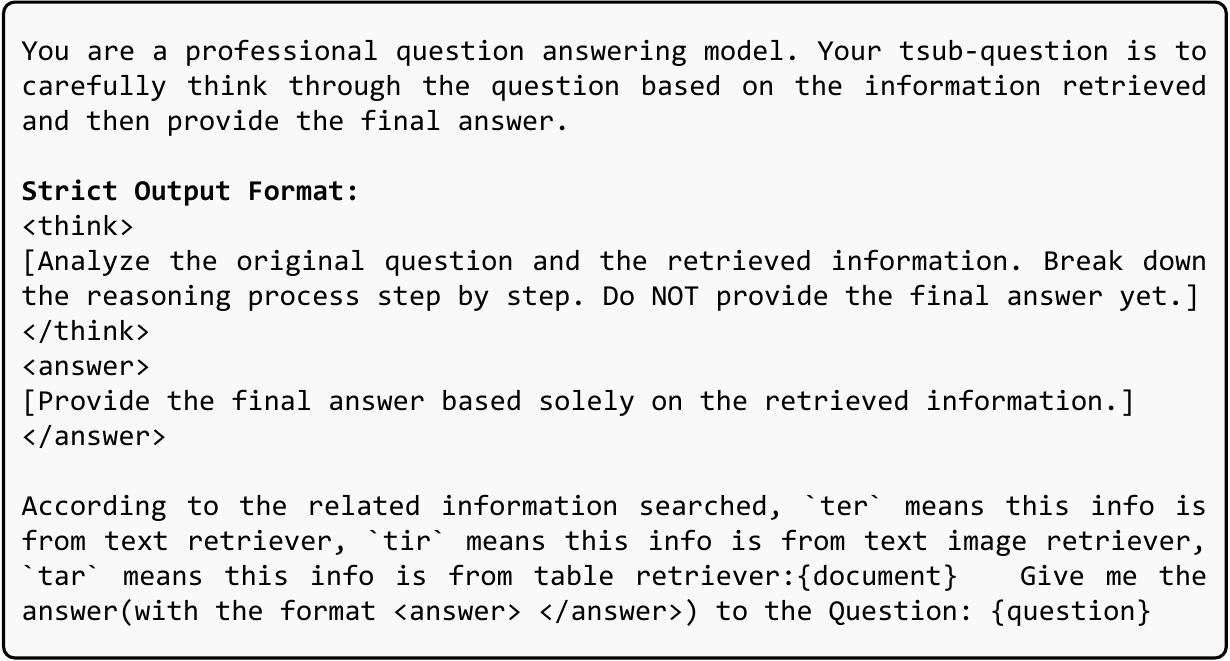}
    \caption{The Prompt Template of \method{} for Answer Generation. We use this prompt template for all \method{} models, including \method{} (SFT), \method{} (Prompt) and \method{} (Step-GRPO).}
    \label{fig:prompt_ans_inter}
\end{figure}
\begin{table*}[t!]
 \centering
\caption{Overall Performance. The best and second-best results are highlighted in bold and underlined, respectively. $*$, $\dagger$, and $\ddagger$ denote statistically significant improvements over CogPlanner, Universal RAG, and IterRetGen w/ all modalities, respectively. \texttt{MMSearch-R1}$^{1}$ adopts the official prompt, which enforces a rigid execution order (accessing the \texttt{Image Expert} followed by the \texttt{Text Expert}), whereas \texttt{MMSearch-R1}$^{2}$ uses our redesigned prompt to encourage flexible expert selection based on query requirements.}
 \label{tab:overall}
 \begin{tabular}{lcccccccc}\toprule
    \multirow{2}{*}{\textbf{Method}} & \multirow{2}{*}{\textbf{ Experts}} & \multicolumn{3}{c}{\textbf{In Distribution}} & \multicolumn{3}{c}{\textbf{Out of Distribution}} & \multirow{2}{*}{\textbf{  Avg.  }}    \\ 
    \cmidrule(lr){3-5}\cmidrule(lr){6-8} 
     & & \textbf{Open-WikiTable} & \textbf{2WikiMQA} & \textbf{ InfoSeek } & \textbf{ Dyn-VQA } & \textbf{ TabFact } & \textbf{ WebQA }\\\midrule
     \rowcolor{gray!20}\multicolumn{9}{l}{\textbf{Vanilla (M)LLMs}} \\\midrule
     Qwen2.5-VL-7B & - & 21.28 & 48.35 & \underline{43.06} & 36.31 & 18.10 & 76.07 & 40.53 \\
     R1-Distill-Qwen-32B & - & 22.75 & 51.78 & 37.20 & \textbf{39.98} & 19.10 & 79.41 & 41.70 \\
     
     \midrule
     \rowcolor{gray!20}\multicolumn{9}{l}{\textbf{Vanilla RAG Methods}} \\\midrule
     Vanilla RAG & Text & 15.38 & 48.20 & 31.88 & 13.26 & 29.20 & 79.04 & 36.16 \\  
     Vanilla RAG &   Image   & 13.77 & 43.95 & 43.03 & 14.03 & 27.50 & 79.76 & 37.01 \\    
     Vanilla RAG & Table & 53.35 & 41.89 & 33.37 & 12.29 & 27.90 & 75.73 & 40.76 \\    
     Vanilla RAG & All & 49.99 & 48.31 & 39.06 & 14.12 & 34.90 & 76.63 & 43.84 \\   
     
     \hdashline
     IRCoT & Text & 5.78 & 24.70 & 16.80 & 16.61 & 5.20 & 46.16 & 19.21 \\
     IRCoT & Image & 5.32 & 18.94 & 27.47 & 18.04 & 2.10 & 47.06 & 19.82  \\
     IRCoT & Table & 35.52 & 11.13 & 17.22 & 14.95 & 8.30 & 44.46 & 21.93 \\
     IRCoT & All & 39.44 & 25.77 & 23.41 & 22.23 & 9.60 & 49.48 & 28.32\\
     
     \hdashline
     IterRetGen & Text & 14.59 & 50.07 & 39.59 & 36.18 & 30.60 & 83.88 & 42.49 \\
     IterRetGen &   Image   & 12.38 & 43.93 & 41.35 & 32.66 & 30.80 & 84.50 & 40.94 \\
     IterRetGen & Table & 36.74 & 42.91 & 40.53 & 32.44 & 36.90 & 83.58 & 45.52\\
     IterRetGen & All & 38.95 & 50.99 & 40.94 & 36.08 & 38.60 & 84.19 & 48.29\\  

     \hdashline
     Search-O1 & Text & 9.72 & 28.12 & 18.52 & 16.40 & 29.60 & 16.78 & 19.86 \\
     \midrule
     \rowcolor{gray!20}\multicolumn{9}{l}{\textbf{RAG Methods w/ Expert Routing}} \\\midrule
     CogPlanner & All & 16.50 & 49.28 & 42.23 & 36.60 & 33.10 & 84.82 & 43.76 \\
     UniversalRAG & All & 31.12 & 47.30 & 37.25 & 11.91 & 26.00 & 79.48 & 38.84 \\
     OmniSearch & All & 7.72 & 31.02 & 24.45 & 18.94 & 2.30 & 58.02 & 23.24\\
     MMSearch-R1$^1$& All &6.50&26.21&13.25&6.86&7.90&2.67&10.57\\
     MMSearch-R1$^2$& All &7.43&19.41&29.09&21.78&39.20&40.75&26.28\\
     \hdashline
     \method-3B & All & \underline{53.85}\rlap{$^{*\dagger\ddagger}$} & \underline{55.10}\rlap{$^{*\dagger\ddagger}$} & 37.45 & 37.58\rlap{$^{*\dagger\ddagger}$} & \textbf{52.60\rlap{$^{*\dagger\ddagger}$}} & \underline{89.54}\rlap{$^{*\dagger\ddagger}$} & \underline{54.37}\rlap{$^{*\dagger\ddagger}$}\\
     \method-7B & All & \textbf{53.90\rlap{$^{*\dagger\ddagger}$}} & \textbf{55.47\rlap{$^{*\dagger\ddagger}$}} & \textbf{43.60\rlap{$^{*\dagger\ddagger}$}} & \underline{39.24}\rlap{$^{*\dagger\ddagger}$} & \underline{52.40}\rlap{$^{*\dagger\ddagger}$} & \textbf{90.92\rlap{$^{*\dagger\ddagger}$}} & \textbf{55.93\rlap{$^{*\dagger\ddagger}$}}\\
    \bottomrule
 \end{tabular}
\end{table*}


\textbf{Implementation Details.} \label{sec:implementationdeatil}
We adopt Qwen2.5-VL-3B and 7B~\cite{bai2025qwen2} as the backbone models for \method{}. 
For constructing the golden trajectories, we employ Qwen2.5-VL-7B as the teacher model for Visual QA, while utilizing R1-Distill-Qwen-32B~\citep{guo2025deepseek} for Text and Table QA.
We set the maximum number of reasoning and retrieval steps $T$ to 3 to balance efficiency and depth. During training via Step-GRPO, we generate $G=8$ rollouts for each query with a temperature of $1.0$. The learning rate is set to $1.0 \times 10^{-6}$ with a cosine decay schedule, and the weight decay is $1.0 \times 10^{-2}$. To ensure stable optimization, we set the clipping parameter $\epsilon=0.2$ and the maximum gradient norm to $1.0$. Figure~\ref{fig:prompt_router} and Figure~\ref{fig:prompt_ans_inter} show the prompt of retrieval and answer process of \method{}

\section{Evaluation Results}\label{sec:evaluate}
In this section, we first present the performance of \method{} across various QA tasks. We then conduct ablation studies to examine the effectiveness of different training strategies. Next, we analyze how \method{} performs retrieval expert planning during the reasoning. Finally, we provide case studies to analyze the behavior of \method{}.

\subsection{Overall Performance}\label{sec:overall}
As shown in Table~\ref{tab:overall}, we compare \method{} with
several baseline models, including vanilla (M)LLMs, RAG methods without expert routing, and RAG methods with expert routing.

Overall, \method{} consistently outperforms all baseline models, achieving an average performance gain of approximately 7\%, which demonstrates its effectiveness. Moreover, \method{} exhibits consistent improvements across all evaluated tasks by adaptively interacting with different retrieval experts to gather task-specific knowledge that satisfies the information needs of MLLMs when answering a given query. These results highlight the strong generalization capability of \method{} and its potential to serve as a universal solution for QA tasks of different modalities.
Moreover, these vanilla RAG models demonstrate that using retrieval experts from a single modality leads to markedly different performance across tasks, confirming the crucial role of modality-specific retrieval experts in answering task-dependent queries. When aggregating information retrieved from all available experts, the resulting models achieve only limited improvements compared to RAG models that rely on a predefined, task-appropriate retrieval expert. This observation indicates that simply concatenating documents from multiple modalities is insufficient for effectively supporting MLLMs in question answering.
Although Search-O1~\citep{li2025search} encourages Large Reasoning Models (LRMs) to perform adaptive retrieval during the reasoning process, it underperforms relative to other approaches. This degradation may stem from the limitations of prompting-based methods in reliably guiding LRMs to invoke and utilize retrieval tools, as such capabilities may not have been adequately acquired during pretraining.
In contrast, \method{}-3B and \method{}-7B achieve significant performance gains over these RAG baselines by explicitly coordinating interactions among the mixture-of-retrieval experts, enabling more effective knowledge acquisition and utilization through collaborative retrieval and reasoning.
Furthermore, \method{} outperforms RAG approaches equipped with expert selection by more than 10\% improvements, highlighting its effectiveness in extending the deep reasoning capabilities of MLLMs for the collaboration of retrieval experts. In contrast to these retrieval expert routing methods, \method{} leverages a Step-GRPO strategy to optimize MLLMs by encouraging richer interactions with retrieval experts, rather than mimicking golden reasoning trajectories. Specifically, it stimulates MLLMs to explore diverse retrieval expert selection outcomes and corresponding answer candidates through sampling, which are then used for reasoning optimization. This strategy enables MLLMs to become more familiar with the behaviors and utilities of different retrieval experts, thereby facilitating more effective and informed expert interactions during the reasoning process.

\begin{table*}[t]
 \centering
 \caption{Ablation Study. All models are implemented with Qwen2.5-VL-7B. $\dagger$ and $\ddagger$ indicate statistically significant improvements over \method{} (Prompt) and \method{} (SFT), respectively.}
 \label{tab:ablationsr}
 \begin{tabular}{lccccccc}
    \toprule
    \multirow{2}{*}{\textbf{Method}} 
    & \multicolumn{3}{c}{\textbf{In Distribution}} 
    & \multicolumn{3}{c}{\textbf{Out of Distribution}} 
    & \multirow{2}{*}{\textbf{Avg.}} \\
    \cmidrule(lr){2-4} \cmidrule(lr){5-7}
    & \textbf{ Open-WikiTable } & \textbf{2WikiMQA} & \textbf{InfoSeek} 
    & \textbf{Dyn-VQA} & \textbf{TabFact} & \textbf{WebQA} & \\


    \midrule
    \rowcolor{gray!20} \multicolumn{8}{l}{\textbf{\method{} (Random Expert)}} \\
    \midrule
    Prompt & 18.88 & 41.44 & 25.70 & 25.16 & 25.90 & 77.79 & 35.81 \\
    SFT & 24.60 & 41.91 & 27.35 & 27.09 & 43.10 & 79.81 & 40.64 \\
    Step-GRPO & 44.03 & 51.08 & 40.56 & 38.18 & 50.50 & 90.66 & 52.50 \\
    \midrule
    \rowcolor{gray!20} \multicolumn{8}{l}{\textbf{\method{} (Self-Routing)}} \\
    \midrule
    Prompt & 23.97 & 41.56 & 24.43 & 26.19 & 25.20 & 77.94 & 36.55 \\
    SFT  & 28.12 & 42.65 & 31.35 & 29.62 & 47.71 & 76.75 & 42.70 \\
    GRPO & 10.41 & 28.56 & 24.18 & 17.23 & \textbf{54.10} & 11.94 & 24.40 \\
     Step-GRPO & 53.95\rlap{$^{\dagger\ddagger}$} & \textbf{55.47\rlap{$^{\dagger\ddagger}$}} & \textbf{43.60\rlap{$^{\dagger\ddagger}$}} & \textbf{39.24\rlap{$^{\dagger\ddagger}$}} & 52.40\rlap{$^{\dagger\ddagger}$} & \textbf{90.92\rlap{$^{\dagger\ddagger}$}} & \textbf{55.93\rlap{$^{\dagger\ddagger}$}} \\
     w/o Expert Selection & \textbf{56.65\rlap{$^{\dagger\ddagger}$}} & 50.26\rlap{$^{\dagger\ddagger}$} & 41.20\rlap{$^{\dagger\ddagger}$} & 38.96\rlap{$^{\dagger\ddagger}$} & 51.80\rlap{$^{\dagger\ddagger}$}& 90.90\rlap{$^{\dagger\ddagger}$} & 54.96\rlap{$^{\dagger\ddagger}$} \\
    \bottomrule
 \end{tabular}
\end{table*}

\subsection{Ablation Studies}\label{sec:ablation}
This section presents ablation studies to evaluate the effectiveness of different training strategies, including Prompt, SFT, GRPO, and Step-GRPO. To examine the contribution of expert routing, we further compare two evaluation settings: (1) random expert, where a retrieval expert is randomly selected whenever a search action is triggered during the reasoning process, and (2) self-routing, where the MLLM autonomously determines which retrieval expert to interact with. In addition, we include Step-GRPO w/o Expert Selection as a comparison, which neglects the expert selection results and instead feeds the relevant documents from all retrieval experts as the context to assist in answering the query.

As shown in Table~\ref{tab:ablationsr}, the prompt-based methods typically yield less than 1\% improvement when adopting the self-routing strategy when compared to the random expert, indicating that merely relying on the internal reasoning capability of MLLMs is insufficient to effectively coordinate multiple retrieval experts when answering a given query. In contrast, the performance of the model trained with SFT is consistently improved over these prompt-based methods, suggesting that supervised fine-tuning better equips MLLMs with expert selection and knowledge utilization capabilities.
By applying the Step-GRPO training strategy, \method{} consistently outperforms these SFT methods by more than 13\%, strongly demonstrating the effectiveness of Step-GRPO in unlocking the potential of MLLMs to cooperate with diverse retrieval experts. This improvement can be attributed to explicitly encouraging MLLMs to interact more frequently and more effectively with retrieval experts during stepwise reinforcement learning optimization.
Crucially, we observe that the outcome-based reward modeling approach, GRPO, leads to severe model collapse. A plausible explanation is that the sparse reward signal fails to properly attribute final task success to specific intermediate routing decisions, preventing the model from learning effective expert interaction strategies.
Moreover, Step-GRPO consistently outperforms the Step-GRPO w/o Expert Selection variant, in which documents from all retrieval experts are directly provided without selective routing. Although this method incorporates more knowledge overall, it inevitably introduces additional noise during answer generation. In contrast, our selective routing mechanism maximizes information density while minimizing irrelevant distractions.
Notably, Step-GRPO achieves more than 3\% improvement over the Random Expert baseline, which is substantially larger than the gains brought by SFT (approximately 2\%). This result further demonstrates that Step-GRPO is particularly effective in enhancing the ability of MLLMs to select appropriate retrieval experts during multi-step reasoning.

\subsection{Effectiveness of Step-GRPO for Retrieval Expert Interaction}
In this subsection, we investigate how the proposed Step-GRPO method enables MLLMs to more effectively interact with diverse retrieval experts.

\begin{figure}[t]
  \centering
  \begin{subfigure}[b]{0.48\linewidth}
    \includegraphics[width=\linewidth]{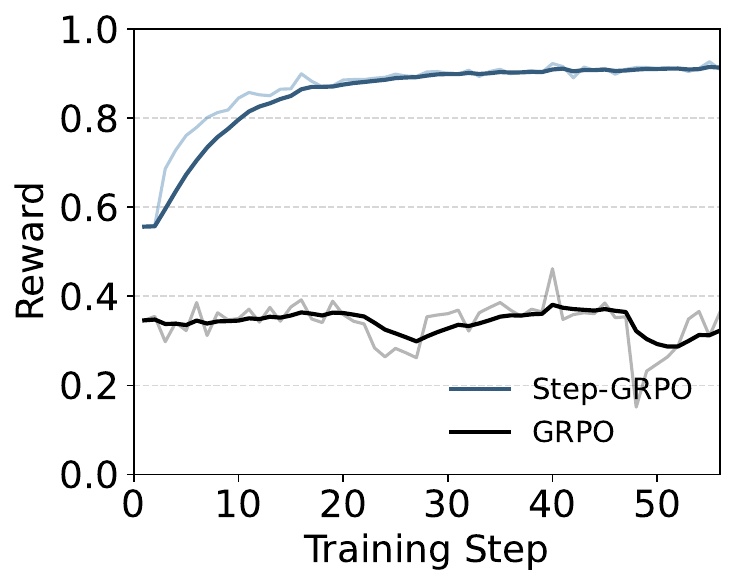}
    \caption{Reward Evolution.}
    \label{fig:reward}
  \end{subfigure}
  \begin{subfigure}[b]{0.48\linewidth}
    \includegraphics[width=\linewidth]{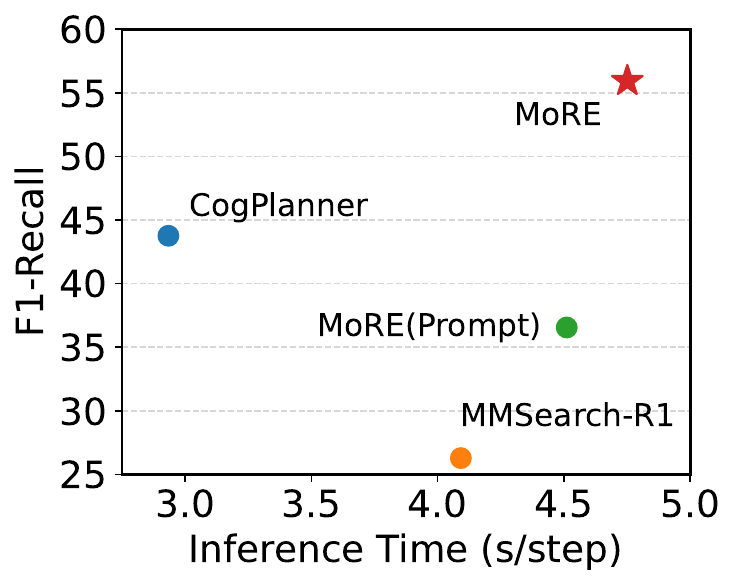}
    \caption{Computational Cost.}
    \label{fig:response}
  \end{subfigure}
  \caption{Training and Inference Effectiveness of Step-GRPO.}
  \label{fig:trainmetric}
\end{figure}

\textbf{Effectiveness of Step-GRPO.}
As illustrated in Figure~\ref{fig:trainmetric}, we first examine the impact of our Step-GRPO training strategy on optimizing the reasoning trajectory.

As shown in Figure~\ref{fig:reward}, training with standard GRPO exhibits significant instability and struggles to converge, which is primarily because relying solely on the final answer for reward makes it difficult for the model to distinguish exactly which intermediate retrieval action contributes to the ultimate success.
In contrast, Step-GRPO delivers stable growth by providing fine-grained interactive feedback from the retrieval experts at each step. This comparison explicitly informs the model of the immediate value of its operations, establishing a clear causal link between routing decisions and information gain.
The effectiveness of this optimization is further corroborated by the efficiency-performance trade-off analysis shown in Figure~\ref{fig:response}.
Compared with simpler expert routing planners, such as CogPlanner~\citep{yu2025unveiling}, \method{} introduces additional computational overhead, but yields more than 10\% performance improvements, indicating that deeper reasoning enables more effective interaction with diverse retrieval experts.
Crucially, when compared with the prompt-based baseline or MMSearch-R1~\citep{wu2025mmsearch}, which incur comparable inference time, \method{} (Step-GRPO) achieves superior QA performance.
These results demonstrate that Step-GRPO does not merely increase computation, but rather teaches the model to make more effective use of its reasoning time to improve solution quality.

        
    

\begin{figure}[t]
  \centering
  \begin{subfigure}[b]{0.49\linewidth}
    \includegraphics[width=\linewidth]{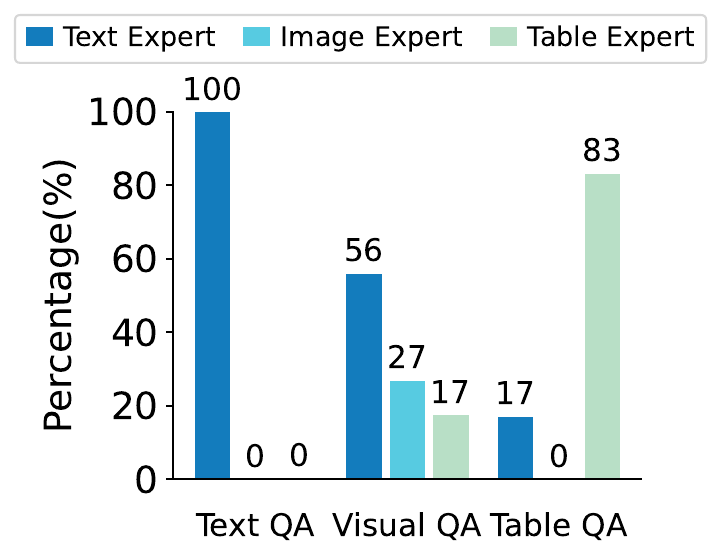}
    \caption{Expert Interaction Distribution of \method{} (Step-GRPO).}
    \label{fig:ret_distribution}
  \end{subfigure}
  \begin{subfigure}[b]{0.49\linewidth}
    \includegraphics[width=\linewidth]{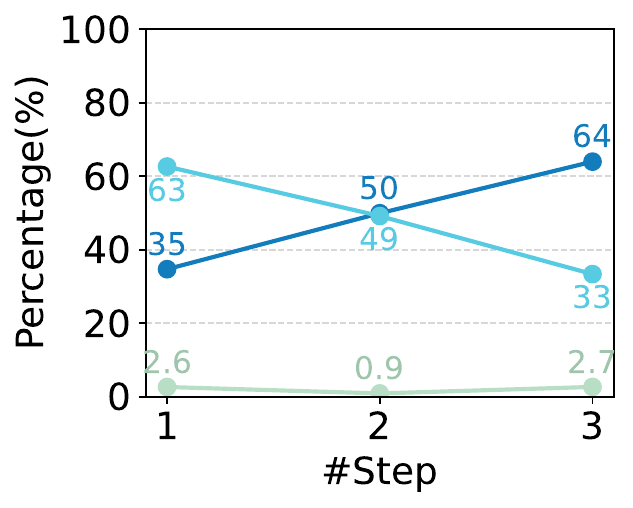}
    \caption{\method{} (Prompt) on VQA.}
    \label{fig:pre_prompt}
  \end{subfigure}
  
  \begin{subfigure}[b]{0.49\linewidth}
    \includegraphics[width=\linewidth]{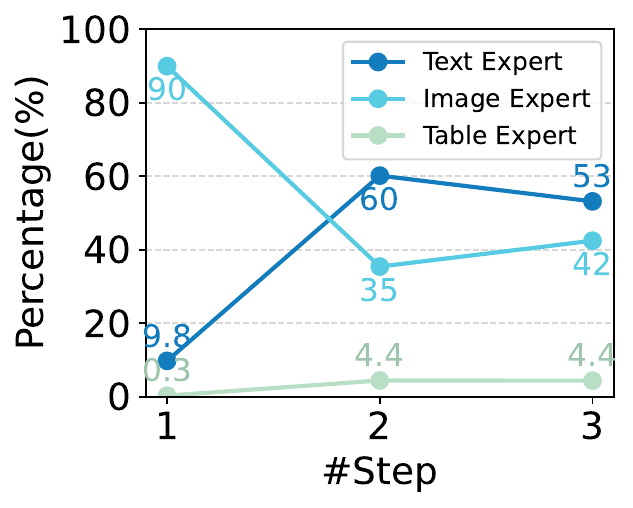}
    \caption{\method{} (SFT) on VQA.}
    \label{fig:pre_sft}
  \end{subfigure}
  \begin{subfigure}[b]{0.49\linewidth}
    \includegraphics[width=\linewidth]{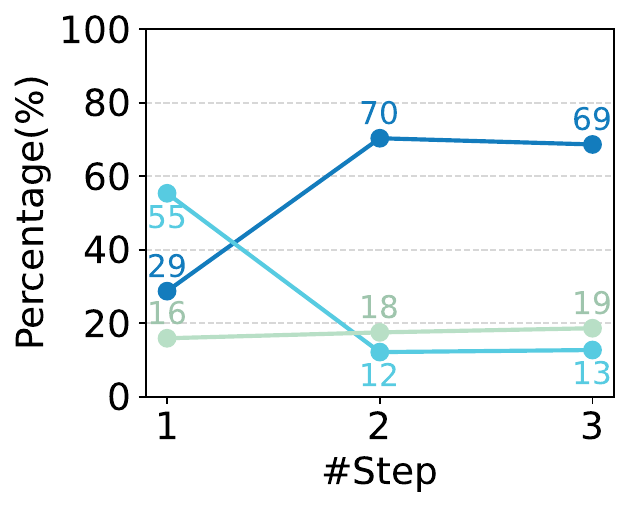}
    \caption{\method{} on VQA.}
    \label{fig:pre_r1}
  \end{subfigure}
  \caption{Expert Interaction Distribution across Reasoning Steps on Different QA Tasks. We compare the evolving retrieval preferences of \method{} under three training strategies: \method{} (Prompt), \method{} (SFT), and \method{} (Step-GRPO). }
  \label{fig:ret_pre}
\end{figure}

\textbf{Expert Interaction of \method{}.}
As shown in Figure~\ref{fig:ret_pre}, we further investigate how \method{} interacts with retrieval experts when handling tasks that require knowledge from different modalities. 

As shown in Figure~\ref{fig:ret_distribution}, we first analyze the expert interaction frequency across different tasks. Overall, \method{} dynamically engages distinct retrieval experts according to task demands, demonstrating its capability to effectively coordinate multiple retrieval experts within a unified QA framework.
For both Text QA and Table QA, \method{} exhibits clear task awareness by predominantly invoking the corresponding task-specific retrieval experts. In the Table QA setting, \method{} occasionally activates the \texttt{Text Expert}, suggesting that certain queries require supplementary textual knowledge to provide contextual grounding for reasoning over tabular data.
In contrast, VQA necessitates interactions with all available retrieval experts, highlighting the intrinsic complexity of visual question answering. Addressing visual queries typically involves a multi-expert interaction pipeline: the \texttt{Image Expert} is used to ground visual entities, the \texttt{Text Expert} retrieves relevant world knowledge, and the \texttt{Table Experts} supply critical complementary information to support holistic reasoning.

To conduct an in-depth analysis of the routing dynamics in VQA scenarios, we examine the behavioral evolution of models optimized under three different training paradigms: \method{} (Prompt), which utilizes few-shot prompting for direct expert routing (Figure~\ref{fig:pre_prompt}); \method{} (SFT), which employs supervised fine-tuning to learn reasoning-based routing patterns (Figure~\ref{fig:pre_sft}); and \method{}, which leverages Step-GRPO to learn routing policies through interactive feedback with experts (Figure~\ref{fig:pre_r1}).
As indicated by the evaluation results, \method{} (Prompt) tends to disregard the \texttt{Table Expert} when answering a given query, which risks missing potentially relevant structured information that could facilitate solving VQA tasks. In contrast, \method{} (SFT) demonstrates markedly rigid retrieval patterns and a pronounced form of expert blindness, manifested as a mechanical ``Image Grounding-then-Text Searching'' paradigm. This behavior is likely attributable to memorized reasoning trajectories acquired during training, whereby the model over-relies on the \texttt{Image Expert} at the initial stage and subsequently engages the \texttt{Text Expert} in a fixed manner, irrespective of the actual query demands.
By comparison, \method{} with Step-GRPO encourages a more adaptive, reasoning-driven interaction pattern among retrieval experts, characterized by more frequent engagement with the \texttt{Table Expert} during the reasoning process, in contrast to the relatively limited interactions observed in both \method{} (Prompt) and \method{} (SFT).
These results further indicate that Step-GRPO enables more effective coordination among retrieval experts, allowing them to collaboratively solve VQA tasks rather than rigidly relying on only a subset of experts.

\begin{figure*}[t]
    \includegraphics[width=\linewidth]{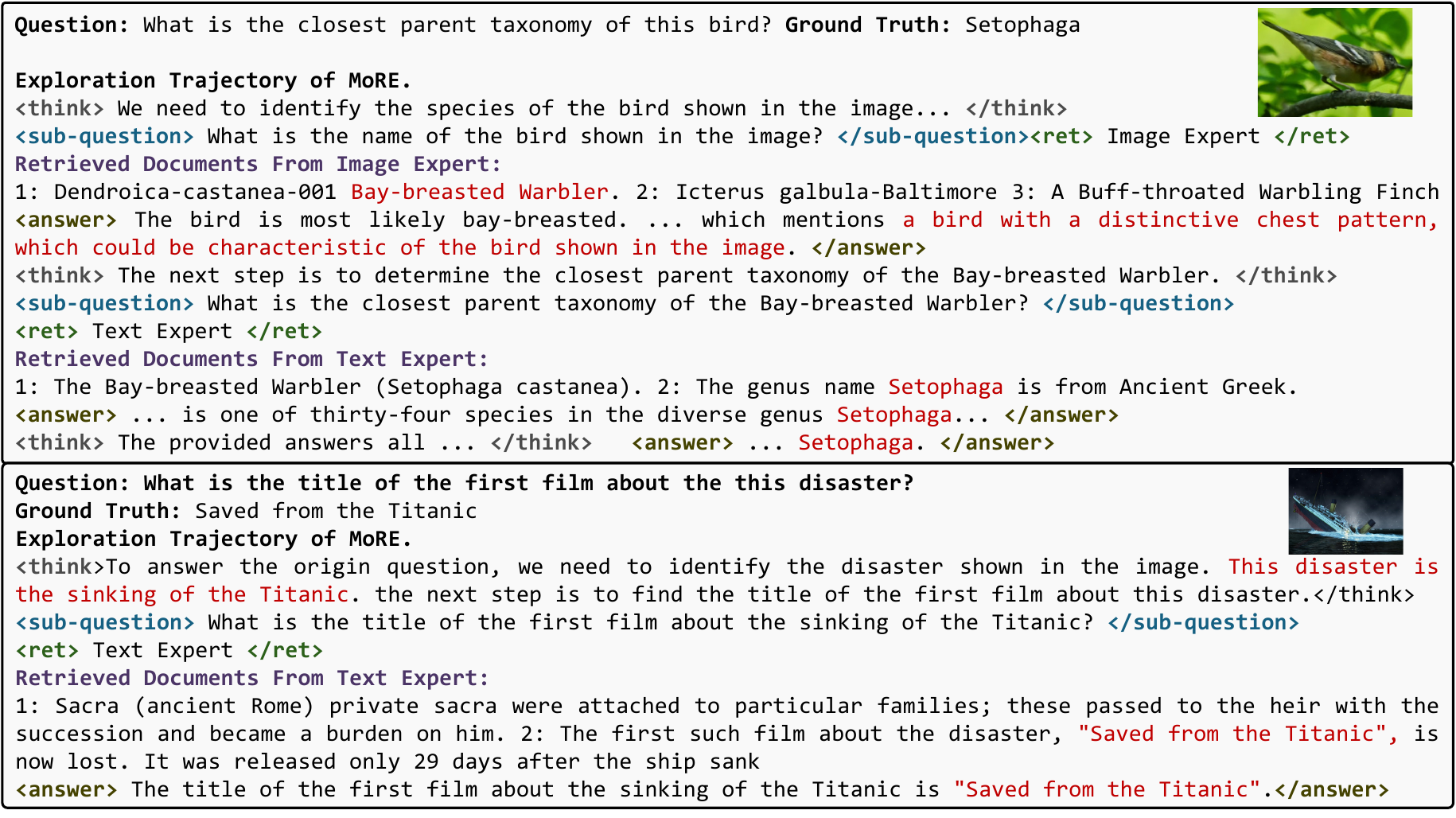}
   \caption{Case Studies. We select two cases from the Visual QA task to illustrate different expert interaction mechanisms. The key information required to answer each question is highlighted in \textcolor[rgb]{0.6902,0.1412,0.0902}{dark red}.}
    \label{fig:case_infoseek}
\end{figure*}

\subsection{Case Study}
As shown in Figure~\ref{fig:case_infoseek}, we sample two cases from the Visual QA task to analyze the expert interaction behavior of \method{}.

In the first case, \method{} is asked to identify the closest parent taxonomy of a bird. To address this query, \method{} first selects the \texttt{Image Expert} to retrieve visually related images that facilitate species identification, and the corresponding captions are retained as contextual evidence. These captions provide informative visual characteristics and relevant entities for subsequent reasoning. Although the retrieved captions may contain noisy or irrelevant information, \method{} performs a reflection step to distill salient visual attributes (e.g., chest patterns), thereby filtering out misleading cues and successfully identifying the bird as a Bay-breasted Warbler. Subsequently, \method{} interacts with the \texttt{Text Expert} to acquire taxonomic knowledge and correctly infers the Bay-breasted Warbler's genus: ``Setophaga''. This reasoning trajectory highlights the effectiveness of \method{} in orchestrating multi-expert interactions to exploit complementary knowledge sources comprehensively.

In the second case, the \method{} model is asked to identify the first film name depicted in the given image. As illustrated in the reasoning process, the model correctly recognizes that the image portrays the sinking of the Titanic. Consequently, \method{} directly routes the query to the \texttt{Text Expert} to retrieve relevant historical knowledge, without invoking the \texttt{Image Expert} to further identify entities within the image. This behavior demonstrates that \method{} is capable of dynamically selecting appropriate experts based on its intermediate reasoning state and knowledge requirements, rather than following a rigid expert interaction pattern. Moreover, this case highlights the importance of encouraging the model to actively interact with retrieval experts during training, instead of overfitting to predefined golden reasoning trajectories.

\section{Conclusion}
In this paper, we propose \method{}, a unified framework for coordinating multiple modality-specific retrieval experts to address queries requiring heterogeneous modality knowledge, enabling MLLMs to dynamically select appropriate experts based on the evolving reasoning context.
Specifically, we introduce Stepwise Group Relative Policy Optimization (Step-GRPO), which overcomes the limitations of sparse outcome-based supervision by exploiting fine-grained interaction signals from retrieval experts, thereby guiding the model to learn effective expert selection policies and to reason about the relative utility of different experts.
Experimental results show that \method{} consistently achieves significant performance improvements across a wide range of benchmarks, while exhibiting stronger adaptability in iterative reasoning scenarios.
By empowering MLLMs to autonomously orchestrate and collaborate with heterogeneous retrieval experts, \method{} represents a meaningful step toward generalizable and agentic RAG systems capable of operating in complex, open-world multimodal environments.

\bibliographystyle{ACM-Reference-Format}
\bibliography{sample-base}

\end{document}